\documentclass[10pt,twocolumn,letterpaper]{article}

\usepackage{iccv}
\usepackage{times}
\usepackage{epsfig}
\usepackage{graphicx}
\usepackage{amsmath}
\usepackage{amssymb}

\usepackage{multirow}
\usepackage[flushleft]{threeparttable}

\usepackage[breaklinks=true,bookmarks=false]{hyperref}

\iccvfinalcopy 

\def\iccvPaperID{8640} 

\ificcvfinal\fi

\begin{document}

\title{KoDF: A Large-scale Korean DeepFake Detection Dataset}

\author{Patrick Kwon\thanks{Equal contribution} \hspace{7pt} Jaeseong You\footnotemark[1] \hspace{7pt} Gyuhyeon Nam \hspace{7pt} Sungwoo Park \hspace{7pt} Gyeongsu Chae\\
MoneyBrain Inc.\\
Seoul, Republic of Korea\\
{\tt\small \{patrick jaeseongyou ngh3053 daniel gc\}@moneybrain.ai}
}

\maketitle
\ificcvfinal\thispagestyle{empty}\fi

\begin{abstract}
   A variety of effective face-swap and face-reenactment methods have been publicized in recent years, democratizing the face synthesis technology to a great extent. Videos generated as such have come to be called deepfakes with a negative connotation, for various social problems they have caused. Facing the emerging threat of deepfakes, we have built the Korean DeepFake Detection Dataset (KoDF), a large-scale collection of synthesized and real videos focused on Korean subjects. In this paper, we provide a detailed description of methods used to construct the dataset, experimentally show the discrepancy between the distributions of KoDF and existing deepfake detection datasets, and underline the importance of using multiple datasets for real-world generalization. KoDF is publicly available at \url{https://moneybrain-research.github.io/kodf} in its entirety (i.e. real clips, synthesized clips, clips with adversarial attack, and metadata).
\end{abstract}

\section{Introduction}

In recent years, the fabrication of facial content in images and videos has become considerably easier and faster, which previously required heavy computing resources and expert knowledge. Latest deep-learning-based technologies have made it possible to handily produce photorealistic fake images and videos by manipulating facial expressions or swapping faces. Soon the word \textit{deepfake} became the de facto term to indicate such facial forgeries synthesized by deep learning models.

\begin{figure}
    \centering
    \includegraphics[width=7.13cm]{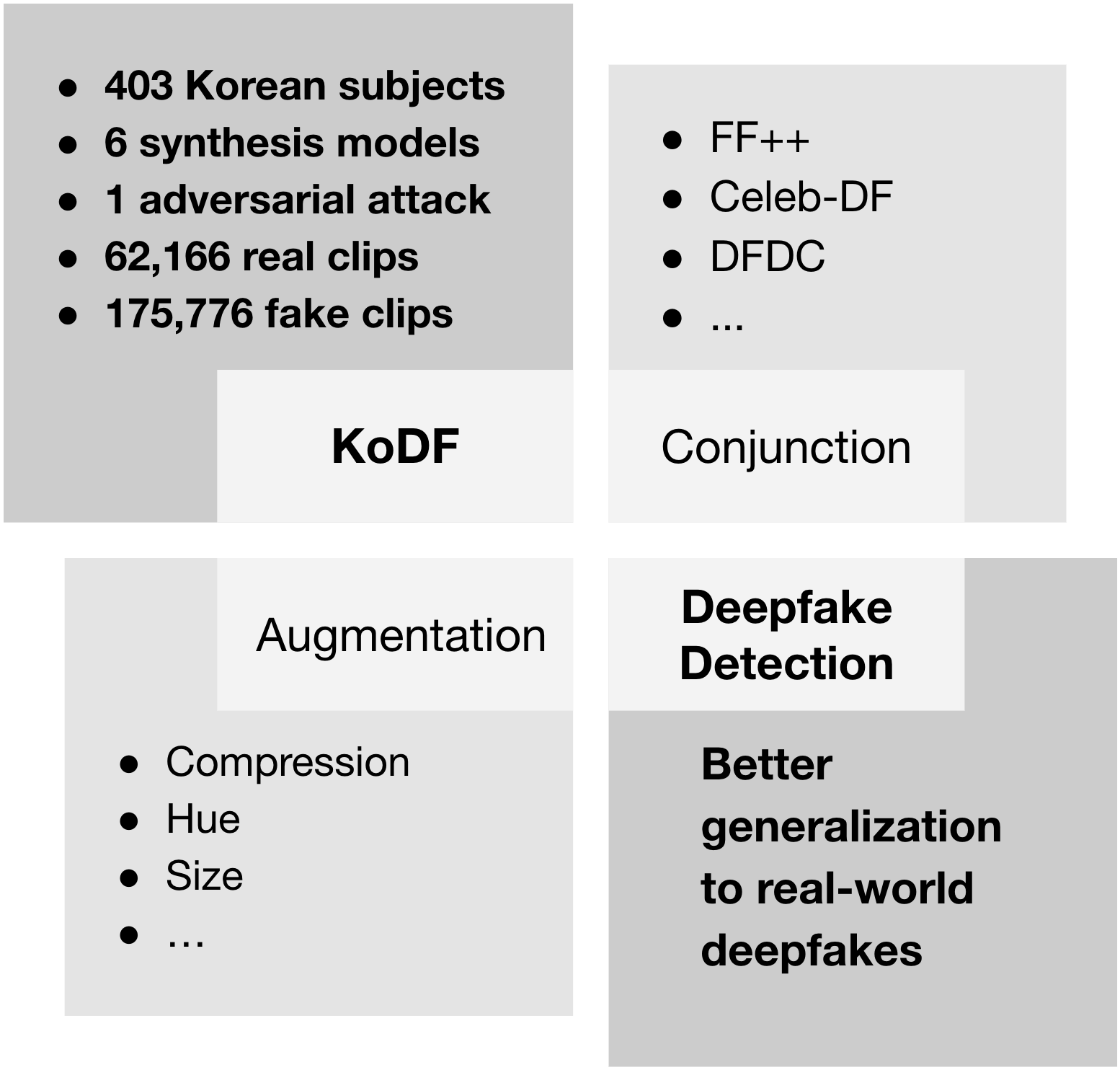} 
    \caption{KoDF is a distribution-controlled large-scale Korean deepfake detection dataset aimed to complement other datasets and to accommodate elaborate augmentation techniques for better generalization to real-world deepfakes.}
    \label{fig:overview}
\end{figure}

While mostly utilized for innocuous purposes such as parody videos \cite{ctrlshiftface} and entertaining apps \cite{AvatarifyInc, Reface}, well-designed deepfakes could be used maliciously to defame an individual \cite{Defame1, Defame2}, propagate disinformation \cite{Disinform1, Disinform2}, or commit fraud \cite{Fraud1, Fraud2}. Due to the growing concerns over deepfakes, there is a recent surge of interest in developing deepfake detection models, and to this end, various public datasets \cite{Dataset_DFTIMIT, Dataset_FFPP, Dataset_CDF, Dataset_GDFD, Dataset_DF1.0, Dataset_VFHQ, Dataset_DFDC} and benchmarks \cite{FF_bench, DFDC_bench, DeeperF_bench, KoDF_bench} have been constructed. They have greatly contributed to encouraging, facilitating, and standardizing deepfake detection research.

In line with these efforts, we release the Korean DeepFake Detection Dataset (KoDF) featuring important differentiations from the previous deepfake detection datasets. KoDF is the largest among publicly available deepfake detection datasets, containing 175,776 fake clips and 62,166 real clips of 403 subjects. The deepfake samples are generated with six different synthesis models. To counterbalance the Asian demographics underrepresented in the existing deepfake detection databases, the participants of KoDF are comprised mostly of Koreans. Finally, the dataset takes various measures to better manage the data distribution regarding the participants' age, sex, and content. Table~\ref{tab:quant} compares KoDF with other public deepfake detection datasets in various aspects.

\begin{table*}[]
\begin{center}
\begin{tabular}{c|rrrcccc}
\hline
Dataset & \begin{tabular}[c]{@{}c@{}}Real\\ videos\end{tabular} & \begin{tabular}[c]{@{}c@{}}Fake\\ videos\end{tabular} & \begin{tabular}[c]{@{}c@{}}Total\\ videos\end{tabular} & \begin{tabular}[c]{@{}c@{}}Rights\\ cleared\end{tabular} & \begin{tabular}[c]{@{}c@{}}Agreeing\\ subjects\end{tabular} & \begin{tabular}[c]{@{}c@{}}Total\\ subjects\end{tabular} & Methods \\ \hline
UADFV               \cite{Dataset_UADFV}   &      49 &      49 &      98 &  No &   0 &      49 & 1       \\
DeepfakeTIMIT       \cite{Dataset_DFTIMIT} &     640 &     320 &     960 &  No &   0 &      32 & 2       \\
FF++                \cite{Dataset_FFPP}    &    1,000 &   4,000 &   5,000 &  No &   0 &      N/A & 4       \\
Celeb-DF            \cite{Dataset_CDF}     &     590 &   5,639 &   6,229 &  No &   0 &      59 & 1       \\
GDFD          \cite{Dataset_GDFD}    &     363 &   3,068 &   3,431 & Yes &  28 &      28 & 5       \\
DF-1.0              \cite{Dataset_DF1.0}   &  50,000 &  10,000 &  60,000 &  No\footnotemark & 100 &     100 & 1       \\
DFDC                \cite{Dataset_DFDC}    &  23,654 & 104,500 & 128,154 & Yes & 960 &     960 & 8       \\ \hline
KoDF                                       &  62,166 & 175,776 & 237,942 & Yes & 403 &     403 & 6 \\ \hline
\end{tabular}
\end{center}
\caption{Quantitative comparison of KoDF to existing public deepfake detection datasets.} 
\label{tab:quant}
\end{table*}

Our contributions are twofold: (1) We propose KoDF that is the largest public deepfake detection dataset, planned and examined for the quality and diversity of its samples. (2) We experimentally demonstrate that none of the established deepfake detection datasets single-handedly suffices in approximating the true deepfake distribution. We then show how utilizing KoDF in conjunction with them for training enhances the generality of a detection model, offering insights into the future strategy in deepfake detection.

\footnotetext{The source videos of DF-1.0 are from 100 paid actors with informed consents, but its 1,000 target videos are taken from FF++, which are collected from YouTube without explicit consent.}

\section{Related Works}

The early deepfake detection databases---the UADFV dataset \cite{Dataset_UADFV} and the DeepfakeTIMIT dataset \cite{Dataset_DFTIMIT}---before FaceForensics++ (FF++) \cite{Dataset_FFPP} and the DeepFake Detection Challenge (DFDC) dataset \cite{Dataset_DFDC} have limitations in quantity and quality. The number of featured identities does not exceed 50, and the total amount of the real and fake videos are less than 1,000. They are collected from untraceable sources, or the agreements from the subjects regarding possible modification and public use of their face are unclear. The fake clips contain a large number of unrealistic synthesized results. Furthermore, the number of employed synthesis methods is only one or two, failing to capture the diversity in the modern means of facial forgery. The two milestone datasets, FF++ and the DFDC dataset, however, overcome many of these difficulties, and other viable deepfake detection databases have been contributed as well.

FF++ \cite{Dataset_FFPP} is the first large-scale dataset to contain 1,000 real videos from YouTube and 4,000 fake videos synthesized by two computer-graphics-based and two learning-based methods. Each of 1,000 raw videos is processed through the four chosen methods, resulting in a total of 5,000 clips. Accompanied is a public leaderboard \cite{FF_bench} where a deepfake detection model can be evaluated against a hidden test set according to benchmark scenario. Before the release of the DFDC dataset, FF++ served as the de facto standard deepfake dataset, thus utilized in various research projects \cite{Agarwal01, Sabir01, Amerini01, Wang01}. However, it fails to fully address some of the aforementioned issues; the size and the diversity of the dataset are still insufficient for optimal training of high-performing neural architectures comprised of huge numbers of parameters, and the subjects' permissions to appear in the database are missing.

In 2020, Amazon Web Services, Facebook, Microsoft, the Partnership on AI’s Media Integrity Steering Committee, and academics have collaborated for DFDC, a large-scale project consisting of a competition, a dataset, and accompanying papers \cite{Dataset_DFDCp, Dataset_DFDC}. The DFDC dataset is released as a part of the challenge. It is thus far the second-largest public deepfake dataset next to KoDF, containing over 960 subjects and more than 120,000 videos. To guarantee the variety of the database, the raw clips are taken from different environmental settings, and the synthesized clips are generated by eight different methods. The challenge was highly successful, encouraging a wide range of researchers to partake in developing effective deepfake detection models, followed by an increased number of research publications on the topic of deepfake \cite{Hernandez-Ortega01, Li01, Masi01, Tolosana01}. However, the DFDC dataset is not without its flaws. Due to the unguided recording process where the participants record themselves, extreme light, audio, and angle conditions are incorporated (e.g. a person talking in a completely dark room), and the data format is inconsistent (e.g. resolution and duration varying across clips). The distribution of participants is not controlled according to age, sex, or race.

Recent years have witnessed other notable public deepfake detection databases varied in focus, composition, and size. Celeb-DF \cite{Dataset_CDF} consists of 590 real videos and 5,639 fake videos. The real videos are taken from YouTube, of which the contents are interviews of 59 celebrities. The fake videos are synthesized by an improved face swap method. The Google DeepFake Detection (GDFD) dataset \cite{Dataset_GDFD} incorporates 3,068 deepfake videos generated based on 363 original videos of 28 consented individuals in 16 different scenes. DeeperForensics-1.0 (DF-1.0) \cite{Dataset_DF1.0} is yet another recent deepfake detection dataset. Its source videos are recordings of 100 paid actors, and the 1,000 target videos are adopted from FF++. 1,000 fake videos are synthesized by swapping each of the source identities onto 10 target videos. Instead of using multiple synthesis methods, it adds diversity utilizing augmentation on both real and fake videos with seven perturbation methods. As a result, 50,000 real and 10,000 fake clips are created, respectively. Albeit these databases are considerably larger and more varied compared to their early correspondences, they have not been utilized across Deepfake detection studies as much as FF++ and the DFDC dataset, so their academic validity is yet to be fully established.

\section{Korean DeepFake Detection Dataset}

\subsection{Contributions}

\subsubsection{Quantity}

KoDF incorporates 62,166 unique 90-second-long real clips (62.8 days) and 175,776 unique deepfake clips of 15 seconds or longer (30.5 days). It surpasses the DFDC dataset, the previously largest public deepfake detection database, in terms of both the total duration (38.4 days of source videos and 12.1 days of generated outputs) and the number of clips (48,190 source videos and 104,500 fake videos). In addition, unlike in the DFDC dataset, audio-swapped or augmented clips do not count as synthesized data points in KoDF. To build the fake portion of the dataset, we resort only to the inference output of the six carefully selected synthesis models (details to be discussed in Section \ref{synth models}) rather than to trivial modifications.

\subsubsection{Controlled Subject Distribution} \label{ctrl_sbj_dst}

KoDF focuses on a situation in which a person talks to a camera, since it is particularly vulnerable to synthetic modifications, and thus frequently targeted by deepfakes. To maximize diversity of the database, we control the 403 participants' distribution according to age, sex, and recording location as shown in Table \ref{tab:demographics}.

\begin{table}[h]
\begin{center}
    {\small
    \begin{tabular}{cccc}
    \hline
    \multicolumn{2}{c}{\multirow{2}{*}{Characteristic}}  & Number  & Percentage (\%) \\ \cline{3-4} 
    \multicolumn{2}{c}{}                                 & 403     & 100             \\ \hline
    \multirow{6}{*}{Age}             & $\sim$19          & 5       & 1.24            \\
                                     & 20$\sim$29        & 205     & 50.87           \\
                                     & 30$\sim$39        & 106     & 26.30           \\
                                     & 40$\sim$49        & 61      & 15.14           \\
                                     & 50$\sim$59        & 19      & 4.71            \\
                                     & 60$\sim$          & 7       & 1.74            \\ \hline
    \multirow{2}{*}{Sex}             & Female            & 205     & 50.87           \\
                                     & Male              & 198     & 49.13           \\ \hline
    \multirow{2}{*}{Location}        & Crowdsourcing     & 353     & 87.59           \\
                                     & Studio            & 50      & 12.41           \\ \hline
    \end{tabular}
    }
\end{center}
\caption{Subject distribution by age, sex, and recording location.}
\label{tab:demographics}
\end{table}

\subsubsection{Quality Assurance and Right Clearance}

KoDF is quality-assured via a meticulous inspection process. Every single one of real and deepfake instances is cross-checked by human eyes and ears for likely issues. The details of the process are provided in Section \ref{real data} and \ref{synth data}. The filtering procedure excludes trivial cases such as a complete failure of synthesis that often appears in other deepfake datasets. KoDF thus includes only true threats where the level of realism is so high that a human cannot easily tell if the clip is real or not.

All the real clips of KoDF are solicited from paid participants. We have informed them of the purpose of the database in great detail, emphasizing the possible consequences where their faces may be manipulated and synthesized. All of them agreed to appear in the database and signed a formal agreement. In addition, all the synthesis models employed have been thoroughly examined for potential license issues. If needed, we have asked for the authors' permission to use their models for the database construction.

\subsubsection{Forward-Looking} \label{fw_look}

Real-world deepfakes would undergo countless modifications (e.g. compression, resizing, manual editing, etc.) in the process of being generated and shared. Elaborate data augmentation is essential to simulate such transformation \cite{Tolosana02}, and consistency in data makes more controlled augmentation possible. We collect and synthesize full HD videos, equalizing the resolution to 1920$\times$1080. Since it is much easier to downgrade a video in quality than to upgrade, the high-resolution clips of KoDF leave a greater and cleaner room for posterior data augmentation. Expecting various augmentation tricks to be applied, we exempt KoDF from any a priori data augmentation (the adversarial attack on 10\% of the data is an addition not a replacement; see section \ref{adv attack}). This is in contrast to the DFDC dataset and DF-1.0 where various perturbations are an inherent part of data. We intend to leave the choice of optimal augmentation techniques to researchers.

Although \textit{face swap} is the best-known method for deepfake creation, deepfake technologies are not simple equivalents of face-swap neural networks. There are a number of other operating means to fake a person's identity in an image or a video. The most concerning one is \textit{face reenactment}, with which one can manipulate actions and expressions of a person in a video, or even in a still image, with an external video or audio source. \cite{Models_FOMM, Models_ATFHP, Models_Wav2Lip}. While being actively researched for obvious commercial use cases, it has not received due attention in the field of deepfake detection. We therefore include an extensive amount of reenactment models' output in our database so that future detection models can be more resilient against reenacted deepfakes.

The Korean subjects (and the eight Southeast Asians) in KoDF counterbalance the DFDC dataset, in which the deficiency of East Asian and Southeast Asian is notable (according to the preview of the database, the proportion of East Asian is  9\%, and that of Southeast Asian is 3\% \cite{Dataset_DFDCp}). The complementary racial composition of KoDF, when put together with other databases, will be critical to build more generalized detection models for real-world applications.

Even in the presence of a working deepfake detection model, the attacking side can take a step further, creating trickier instances to deceive the detector. For example, destructive means to confuse classification models have been devised \cite{AdversarialGoodfellow, AdversarialGaneshan, AdversarialGandhi}. Therefore, we add to KoDF adversarial examples to encourage the development of detection models that are robust against such attacks. Refer to Section \ref{adv attack} for more details.

\subsection{Real Data} \label{real data}

Unlike previous deepfake detection databases that are made of found clips \cite{Dataset_DFTIMIT, Dataset_FFPP}, the source videos of KoDF are recorded specially to constitute the database. By governing the recording process ourselves, we prevent defective instances and control the distribution with regard to recording environment, emotive content, and speech corpus.

Among the 403 subjects, 353 partake in the \textit{crowdsourcing} task where a subject is asked to film oneself for 150 clips, each of which should last over 90 seconds. The first recording of the 150 clips is an \textit{idle} clip where a subject remains in a natural pose saying nothing. In the half of the remaining 149 recordings, a subject reads an assigned script consisting of 10 sentences. These are \textit{script} clips. The remaining 74 clips are \textit{scenario} clips, in which a subject chooses or makes up a question, and provides his or her responses in a given time. To add more diversity, we introduce minor variations in terms of camera angle, focal length, recording location, background, composition of props, and lighting.

Each clip belongs to one of the three emotive categories: \textit{positive}, \textit{negative}, and \textit{neutral}. The sentences of the script clips and the questions of the scenario clips are formulated as such per recording. The purpose of this task design is to facilitate the recording process for the subjects who are mostly amateurs with no experience in shooting footage of themselves. The aforementioned tactics help vary the subjects’ responses while providing them enough materials to continue talking without prolonged pauses.

The sentence corpus for the script clips is comprised of definitions and examples crawled from Standard Korean Language Dictionary \cite{KoreanDict}. The found sentences are screened if the length is too short or long, or if non-Korean symbols are included. They are subsequently organized by sentence type: \textit{statement}, \textit{question}, and \textit{exclamation}. We adjust the occurrence ratio between the statement type and the other two types in a script clip to 8:2 for the diversity of expression. The chosen sentences are machine-tagged with respect to their emotion category based on Kunsan National University Korean Sentiment Lexicon \cite{KunsanSentiment} by simply accumulating the valence score per token. A script clip contains 10 sentences of the same emotion category; for example, for a positive clip, 10 positive sentences are assigned.

For the scenario clips, 420 questions are collected from amateur writers and proofread. The questions are evaluated for their emotive quality by three annotators and then categorized by the majority rule. In each clip, a subject is asked to choose a question of the corresponding category and answer it during the given time. While a subject is free to make up his or her own question instead of choosing from the question bank, the made-up question needs to meet the emotive category in terms of the content of the question and the expected answer. The purpose of the scenario design is to relax the amateur participants and to widen the range of speaking style, complementing the relatively rigid dynamics and monotonous prosody of the script clips.

The remaining 50 subjects participate in the \textit{studio} task, where the recording environment and the task design differ from those for the crowdsourcing task. They record high-quality video footage at a professional studio with a skilled director against a green screen. A subject carries out eight recording runs in one or two sessions on different days. He or she reads 300 sentences per run, and each of the runs takes approximately 35 minutes. The eight long recordings are later split into 90-second intervals, resulting in 184 clips in total. These clips are equivalent to the script clips in the crowdsourcing task, and we do not include scenario design for the studio task.

The collected real clips are manually inspected for various possible defects: (1) audio-video sync problem, (2) excessive background noise, (3) utterances severely hindered or stuttered, (4) extreme lighting conditions, and (5) face located far outside of the central region. If any of these problems is detected during the checking process, the subject is requested to shoot the corresponding clip again.

\subsection{Synthesized Data} \label{synth data}

We employ six different models to generate deepfake clips. Among them, FaceSwap \cite{Models_DFFS}, DeepFaceLab \cite{Models_DFL}, and FSGAN \cite{Models_FSGAN} are face swapping models. First Order Motion Model (FOMM) \cite{Models_FOMM} is a \textit{video-driven} face-reenactment model. The remaining two, Audio-driven Talking Face Head Pose (ATFHP) \cite{Models_ATFHP} and Wav2Lip \cite{Models_Wav2Lip}, are \textit{audio-driven} face-reenactment models. Before and after the actual synthesis, video clips are processed to reduce artifacts and enhance fidelity.

Hereafter, the terms \textit{target} and \textit{source} are used to denote different facial identities in face swapping; the target is the to-be-replaced facial content of the base video, and the source is the facial content to replace the target. Thus a clip resulting from a face swapping method looks identical to the target video, except for the facial identity which is of the source. On the other hand, the source and the target are one and the same in face-reenactment methods; their goal is to manipulate the pose or expression of the source person while maintaining the rest. The driving audio or video can be taken from different identities.

\subsubsection{Preprocessing}

All the raw clips are preliminarily processed subject-wise. We employ a facial landmark algorithm 2DFAN \cite{Bulat01}, by which the facial regions are cropped, aligned, and resized to 512$\times$512 pixels. These regions are computed by an affine transformation of positions centered around eyes and nose. To train face swapping models, 4,000 to 5,000 face frames are selected according to their sharpness and to the diversity of face angles (reenactment models require little or no additional training).

\subsubsection{Synthesis Models} \label{synth models}

\begin{figure}
    \centering
    \includegraphics[width=7.0cm]{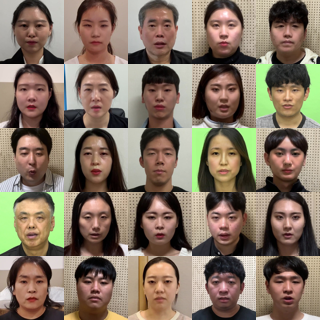}
    \caption{A selection of synthesized frames. Each row is created using FaceSwap, DeepFaceLab, FSGAN, FOMM, and Wav2Lip from top to bottom.}
    \label{fig:sample}
\end{figure}

\begin{figure}
    \centering
    \includegraphics[width=8.4cm]{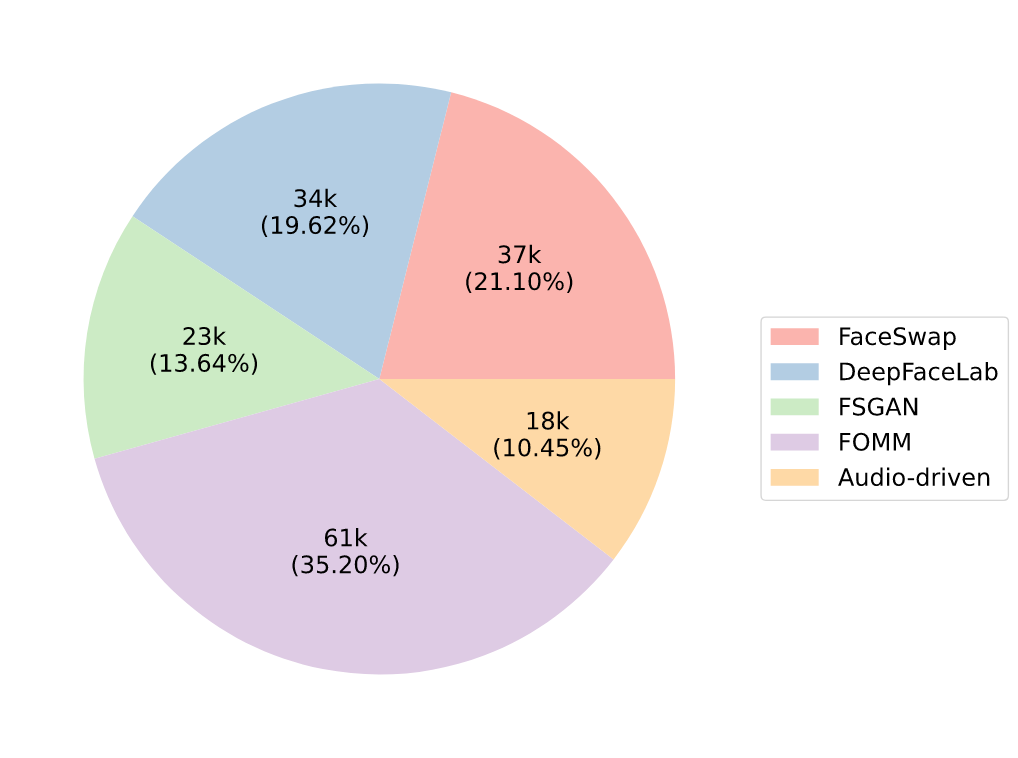}
    \caption{KoDF video distribution by synthesis methods. Audio-driven includes ATFHP and Wav2Lip.}
    \label{fig:pie_chart}
\end{figure}

The synthesis models of KoDF are a diverse collection of facial manipulation techniques. Although we try to maintain equal distribution amongst the models, since all of the generated clips are validated under manual screening processes, the number of videos per method is not equal. Some methods guarantee a stable level of realism, accounting for a larger chunk while others respond sensitively to lighting and noise, resulting in a number of unusable clips that do not make it into KoDF. Figure \ref{fig:sample} shows the example frames generated by the selected methods, and the distribution of the synthesized videos is illustrated in Figure \ref{fig:pie_chart}.

\textbf{FaceSwap} FaceSwap \cite{Models_DFFS} is an open-source face-swapping software initially developed by a Reddit user \textit{/u/deepfakes} and later maintained by its developing community. It bears an encoder-decoder architecture. Two separate decoders handle the source and target faces respectively while sharing one encoder. The three networks are trained concurrently; the encoder learns the non-identity features, while the two decoders capture traits that are more contingent to each of the identities. The model consequently infers face images that maintain the source identity while matching the target’s non-identity features. We include this model because of its historical importance as the starting point where the term deepfake originates from.

\textbf{DeepFaceLab} DeepFaceLab \cite{Models_DFL}, currently the most popular means to generate deepfake videos, provides an imperative and easy-to-use pipeline, along with a collection of synthesis models \cite{DeepFaceLabGit}. These models are highly FaceSwap-like, with respect to the convolutional autoencoder architecture as well as the training and inference processes. One of notable improvements from FaceSwap is, however, the intervening network in between the shared encoder and the two decoders. This modification helps capture common non-identity features underlying both the source person and the target person, contributing to robust mapping between the two. In addition, a mixed loss combines structural dissimilarity index with mean squared error, leading to improved fidelity.

\textbf{FSGAN} FSGAN \cite{Models_FSGAN} is capable of both face swapping and reenactment. The model first reenacts the source identity according to the target's pose and expression, and segments the facial regions of both faces. Then it inpaints the missing parts of the reenacted face and blends the completed face with the target, creating the final result. During reenactment, the model selects multiple source frames that correspond most to the target by Delaunay Triangulation, and uses the weighted average of the reenactment results according to barycentric coordinates. This process makes the model subject agnostic so it does not require heavy tuning on every new source. For this paper, we use the face swapping scheme of the official implementation \cite{FSGANGit}. We adopt the author's recommended swap method for better quality of synthesis\footnote{\url{https://github.com/YuvalNirkin/fsgan/wiki/Face-Swapping-Inference}}, which fine-tunes the model for each source-target pair for 800 iterations and utilizes the target mouth area for better teeth quality.

\textbf{FOMM} FOMM \cite{Models_FOMM} is a self supervised network that applies the motion of a driving video sequence to an image where both contain objects of the same category (e.g. faces). It decouples appearance and motion by modeling the movement around the keypoints using affine transformations in a self-supervised manner. It warps the source image based on the motion of the driving video then recovers the warping artifacts by inpainting. We use the official code and the pretrained model \cite{FOMMGit} on VoxCelab dataset \cite{Nagrani01}. This model is chosen to represent the face-reenactment strategy and for its real-world applications. For instance, Open Avatarify \cite{Avatarify}, a popular real-time reenactment tool for video chats, adopts this model.

\textbf{ATFHP and Wav2Lip} We employ two different audio-driven face synthesis models for KoDF: ATFHP \cite{Models_ATFHP} and Wav2Lip \cite{Chung01}. The former reflects the active research domain of face synthesis based on 3D morphable model. ATFHP takes audio and video inputs to create an output video, which retains the identity of the input video while synchronizing facial expressions to the audio. This is achieved by creating a list of 3D model parameters from the input audio and render them to synthesized frames. A memory-augmented GAN module then refines the rendered frames into realistic ones with smooth background transition for various face identities. After pretrained with the Lip Reading in the Wild dataset \cite{Chung02}, it only requires fine-tuning with a small number of frames to learn personalized talking behavior.

Soon after starting to generate audio-driven face-reenactment instances for KoDF using ATFHP, we experiment with a newly published alternative, Wav2Lip. Unlike similar models that generate a talking face from a driving audio with a GAN-based architecture, Wav2Lip utilizes a pretrained lip-sync discriminator, which helps the model to learn the appropriate lip motion according to the audio. To capture the temporal context of speech, the model uses five consecutive face frames and the respective speech content as input.

While the synthesized results from ATFHP are promising, we switch from ATFHP to Wav2Lip due to the relative efficiency of the synthesis process of the latter. For every input identity, the pretrained model of ATFHP \cite{ATFHPGit} requires careful fine-tuning to acquire high-fidelity syntheses that meet our quality assurance criterion. On the other hand, that of Wav2Lip \cite{Wav2LipGit} can efficiently generate samples of proper quality with respect to unseen facial identities without fine-tuning. As a result, 455(2.5\%) and 17,915 (97.5\%) clips are synthesized using ATFHP  and Wav2Lip, respectively.

\subsubsection{Postprocessing}

All the methods listed above produce a sequence of image frames matched to the facial region cropped during the preprocessing step. Because most models fail to reconstruct accurate details around the facial boundaries, necessitated is the process of blending the synthesized outcome back into the original frame.

Using the same facial landmark detection \cite{Bulat01} from the preprocessing stage, we create a facial mask from the synthesized image frame. The border of the mask region goes under a Gaussian blurring process to reduce the artifacts, and the blurred images are blended into the original video frames of corresponding temporal positions. This postprocessing procedure reduces jitters while preserving details around the facial borders.

\subsubsection{Quality Evaluation of KoDF}

Once videos are generated from the synthesis models, they go through a manual screening process where they are presented to two raters and subjected to two questions: (1) \textit{Is the clip of high quality?} and (2) \textit{Can the figure in the clip pass as a real human?} We keep only the clips that win both raters’ approvals for both of these questions. Each clip is presented once in the size of the horizontal cell phone layout. Raters also examine various technical issues regarding orientation, audio-video synchronization, duration, among others. Clips that fail to pass the screening are simply discarded.

The quality of the synthesized output is evaluated with peak signal-to-noise ratio (PSNR), structural similarity index measure (SSIM), Fréchet Inception distance (FID) \cite{FID01}, and average keypoint distance (AKD).

\begin{equation}
PSNR(r,g) = 10 \cdot log_{10}(\frac{MAX_{I}^2}{MSE(r,g)})
\end{equation}
\begin{equation}
SSIM(r,g) = \frac{(2\mu_{r}\mu_{g} + c_{1})(2\sigma_{rg} + c_{2})}{(\mu_{r}^{2} + \mu_{g}^{2} + c_{1})(\sigma_{r}^{2} + \sigma_{g}^{2} + c_{2})}
\end{equation} 
\begin{equation}
AKD(r,g) = \frac{1}{P}\sqrt{\sum_{p=1}^{P}(r_{p} - g_{p})^2}
\end{equation}

\begin{table}[]
    {\small
    \begin{tabular}{c|cccc}
    \hline
    Method          & PSNR $\uparrow$ & SSIM $\uparrow$  & FID $\downarrow$ & AKD $\downarrow$ \\ \hline
    FaceSwap        & 22.10{\scriptsize±2.02} & 0.76{\scriptsize±0.05} & 1.11{\scriptsize±0.08} & 0.21{\scriptsize±0.04} \\
    \footnotesize{DeepFaceLab}     & 21.86{\scriptsize±1.82} & 0.75{\scriptsize±0.05} & 1.12{\scriptsize±0.09} & 0.22{\scriptsize±0.04} \\
    FSGAN           & 21.09{\scriptsize±2.07} & 0.79{\scriptsize±0.08} & 1.07{\scriptsize±0.09} & 0.16{\scriptsize±0.02} \\
    FOMM            & 26.16{\scriptsize±2.96} & 0.87{\scriptsize±0.04} & 1.00{\scriptsize±0.07} & 0.15{\scriptsize±0.02} \\
    \footnotesize{Audio-driven}    & 24.47{\scriptsize±2.39} & 0.84{\scriptsize±0.06} & 1.09{\scriptsize±0.08} & 0.17{\scriptsize±0.02} \\ \hline
    Total           & 23.72{\scriptsize±3.17} & 0.81{\scriptsize±0.07} & 1.06{\scriptsize±0.09} & 0.18{\scriptsize±0.04} \\ \hline
    \end{tabular}
    }
    \caption{Quality evaluation of KoDF for each synthesis method. Audio-driven includes ATFHP and Wav2Lip.}
    \label{tab:qeval}
\end{table}

\begin{table}[h]
\begin{center}
    {\small
    \begin{tabular}{c|cccc}
    \hline
        Dataset & PSNR $\uparrow$ & SSIM $\uparrow$ & FID $\downarrow$ & AKD $\downarrow$ \\ \hline
        FF++    & 23.11\footnotesize{±3.22} & 0.77\footnotesize{±0.09} & 1.11\footnotesize{±0.08} & 0.26\footnotesize{±0.08} \\
        DFDC    & 24.54\footnotesize{±3.23} & 0.79\footnotesize{±0.08} & 1.14\footnotesize{±0.09} & 0.25\footnotesize{±0.13} \\
        DF-1.0  & 22.15\footnotesize{±1.76} & 0.76\footnotesize{±0.06} & 1.11\footnotesize{±0.06} & 0.19\footnotesize{±0.11} \\ \hline
        KoDF    & 23.72\footnotesize{±3.17} & 0.81\footnotesize{±0.07} & 1.06\footnotesize{±0.09} & 0.18\footnotesize{±0.04} \\ \hline
    \end{tabular}
    }
    \caption{Quality comparison of KoDF and other datasets.}
    \label{tab:qcomp}
\end{center}
\end{table}

For the evaluation, we randomly choose 500 real clips and 500 corresponding synthesized clips. From each of the fake samples, 100 frames are uniformly extracted, and their real matches are taken from the identical temporal positions. Each metric is averaged for these 100 pairs to compute the value for each synthesized clip. Table \ref{tab:qeval} shows the results by the synthesis methods of KoDF, and Table \ref{tab:qcomp} compares KoDF with the FF++, DFDC, and DF-1.0 datasets.\footnotemark

\footnotetext{GDFD is not included here due to the time differences between its corresponding real and fake clips and the lack of metadata to correct them.} 

\subsection{Adversarial Attack} \label{adv attack}

In the midst of the burgeoning interest in deepfake detection technologies, researchers are thinking ahead to the next step: adversarial attacks to fool detecting models. Methods that are known to be disruptive for deepfake detection include gradient-based adversarial attacks \cite{AdversarialGandhi} and intentional noises that hide spatial and spectral artifacts arising from a synthesis model \cite{Huang01}. There is even a toolbox to create such adversarial instances \cite{Goodman01}, and researchers report the vulnerability of deepfake detection models against black-box attacks \cite{Neekhara01, Hussain01}.

The fast gradient sign method \cite{AdversarialGoodfellow} is the means chosen for KoDF to simulate potential malicious attempts to evade detection, for it is the most widely known adversarial method. The process of creating adversarial examples are as follows: we train a preliminary detector model, obtain a sequence of noise frames that confuse the preliminary model, and mix the noise sequence with the corresponding input clip. This strategy is based on the assumption that most detectors are similarly structured as the preliminary detector, which allows the generated adversarial examples to generalize their elusive effects across architectural details of detectors.

We build the preliminary model by topping a pretrained EfficientNet-B4 \cite{pmlr-v97-tan19a} base with two fully connected layers. The model is then trained with a portion of KoDF, which is structured as follows: from 10\% of randomly selected fake samples, frames between the 150th and 450th positions are extracted at a chance of 0.8\%, and from 2\% of real clips, one in a hundred frames is stochastically drawn. Approximately 40,000 frames are compiled for each category, and the model is trained for 10 epochs.

To create noised instances, every one in ten clips is randomly selected regardless of its real or fake category and decomposed into a sequence of frames. Each frame is iteratively fed into the trained preliminary model. Obtained is the sign of the generated gradients with respect to the input image $x$, which is scaled and added to the original frame. 

\begin{equation}
x_{adv} = x + \epsilon \cdot \textrm{sign}(\nabla_{x}J(\theta,x,y))
\end{equation} 

The scaling factor $\epsilon$ for the noise is varied between 0.01, 0.05, and 0.1, and the preliminary model is retrained three times over the course of time, to generate noises of varying nature and intensity. The noised frames are once again put together into a video with the addition of audio from the original source.

\section{Detection Evaluation} \label{detection evaluation}

\begin{figure*}
    \centering
    \includegraphics[width=17.3cm]{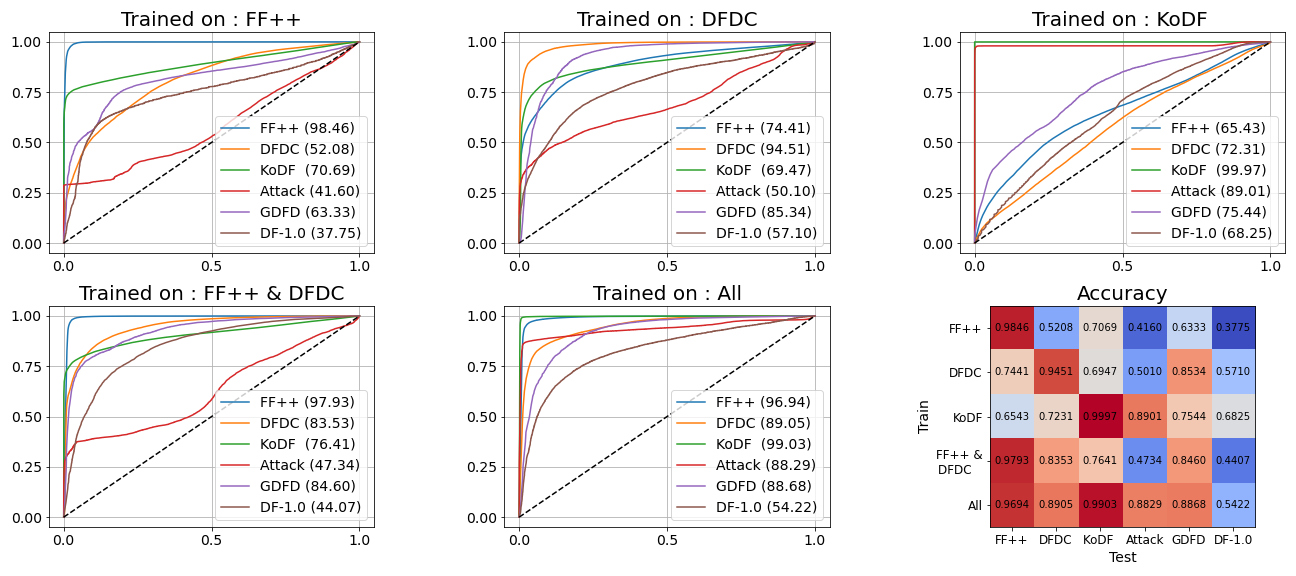}
    \caption{ROC curves of the DFDC winning detection model trained on FF++, the DFDC dataset, KoDF, and their combinations. All training set variants are of equal size. The trained models are evaluated on unseen test sets, including the adversarial samples of KoDF (Attack), the GDFD dataset, and DF-1.0. We also include a corresponding accuracy heat map.}
    \label{fig:roc}
\end{figure*}

The ultimate goal of a deepfake detection dataset would be to help develop a general detection model that performs well against a variety of real-world deepfake cases. Most studies on deepfake detection are designed so as to measure how their proposed detection models perform based on a certain deepfake detection dataset. The premise here is that the target deepfake detection dataset is a good approximation of the distribution of real-world deepfake instances.

In the subsequent experiments, we investigate if existing deepfake detection datasets guarantee a sufficient level of generality and how they fare when mixed and tested with out-of-domain data. To this end, we train the winning model of the DFDC competition \cite{selim} with combinations of the FF++, DFDC, and KoDF datasets. The multiple variants are then evaluated against unseen test sets, which include the adversarial samples of KoDF (Attack), the GDFD dataset, and DF-1.0.

For fair comparison, the DFDC dataset and KoDF are sampled to match the size of FF++. 1,000 real videos and 4,000 fake videos are randomly chosen from the two datasets. Following the preprocessing method of \cite{selim}, we use MTCNN \cite{MTCNN} to extract a face from every frame of the FF++ and DFDC clips, and frames with unrecognizable faces are ignored. In the case of KoDF, we randomly select 300 frames from each clip to match the total number of face samples per dataset. The extracted facial frames are then divided into training, validation, and test sets for each dataset, with a ratio of 8:1:1. We train the DFDC winning model on different combinations of the three training sets, all of which are of equal size. After 30 epochs of training, the epoch with the minimum loss value on the validation set is chosen, and the trained models are evaluated on each of the six test sets. Figure \ref{fig:roc} summarizes the results.

The results show that each of the three deepfake detection datasets, on its own, is not enough to approximate the true distribution of deefakes in the wild, resulting in detection models that are incapable of adapting to different deepfake detection datasets. The issue is relatively severer with KoDF, but it is only natural since the distribution of our dataset is systematically controlled to a greater extent (as detailed in Section \ref{ctrl_sbj_dst} and \ref{fw_look}). The point here is that, when trained on their combinations, the model becomes much more robust against various out-of-domain data. Noteworthy is the performance improvement when the model utilizes all three in comparison to training with only FF++ and the DFDC dataset. This observation supports the complementary utility of KoDF.

From the experimental results, we can deduce that the deepfake detection task is strongly prone to overfitting, much more so than regular image classification tasks where models learn diverse signals recurring naturally (i.e. local patterns and global structures). On the other hand, deepfake detection models focus on artifacts arising during the generation process, which inevitably vary depending on the synthesis methodologies. An ideal deepfake detection dataset should thus incorporate examples of a maximal variety of deepfake methods and a wide range of real videos. No standalone deepfake dataset published so far achieves sufficient generality to meet these conditions on its own, and a practical solution is to utilize multiple datasets adjoined.

\section{Conclusions}

We present a new large-scale dataset to help researchers develop and evaluate deepfake detection methods. KoDF focuses on Korean subjects to compensate for the Asian underrepresentation of other major deepfake detection databases. It expands the range of employed deepfake methods, regulates the quality of the real clips and the synthesized clips, manages the distribution of subjects according to age, sex, and content, and simulates possible adversarial attacks. While KoDF is an extensive database, our expectation is that it will work even more effectively in the mutual complementation of existing and future deepfake detection databases, including the two milestone datasets FF++ and DFDC. We experimentally demonstrate the benefit of compositing datasets for in-the-wild deepfake detection. We hope KoDF to serve as a stepping stone for future studies in the field of deepfake detection.

\section{Acknowledgements}

We gratefully acknowledge that KoDF was built as part of the AI Training Data Construction Project 2020 hosted by the Ministry of Science and ICT (MSIT) and supported by the National Information Society Agency (NIA) of South Korea. This research was partly supported by the Institute of Information \& Communications Technology Planning \& Evaluation (IITP) grant funded by MSIT (2021-0-00888). 

{\small
\bibliographystyle{ieee_fullname}
\bibliography{egbib}

\begin{thebibliography}{10}\itemsep=-1pt

\bibitem{ATFHPGit}
Audio-driven talking face head pose official implementation.
\newblock \url{https://github.com/yiranran/Audio-driven-TalkingFace-HeadPose}.
\newblock Accessed: 2020-07-16.

\bibitem{AvatarifyInc}
Avatarify, inc.
\newblock \url{https://avatarify.ai/}.
\newblock Accessed: 2020-10-17.

\bibitem{ctrlshiftface}
Ctrl shift face.
\newblock \url{https://www.youtube.com/channel/UCKpH0CKltc73e4wh0_pgL3g}.
\newblock Accessed: 2020-09-27.

\bibitem{DeeperF_bench}
Deeperforensics challenge 2020.
\newblock \url{https://competitions.codalab.org/competitions/25228}.
\newblock Accessed: 2020-08-03.

\bibitem{DeepFaceLabGit}
Deepfacelab.
\newblock \url{https://github.com/iperov/DeepFaceLab}.
\newblock Accessed: 2020-07-16.

\bibitem{DFDC_bench}
Deepfake detection challenge.
\newblock \url{https://www.kaggle.com/c/deepfake-detection-challenge}.
\newblock Accessed: 2020-05-16.

\bibitem{KoDF_bench}
Deepfake video detecting ai competition (using kodf preview dataset).
\newblock \url{https://dacon.io/competitions/open/235655/overview/}.
\newblock Accessed: 2020-12-20.

\bibitem{FF_bench}
Faceforensics benchmark.
\newblock \url{http://kaldir.vc.in.tum.de/faceforensics_benchmark/}.
\newblock Accessed: 2020-07-16.

\bibitem{Models_DFFS}
Faceswap.
\newblock \url{https://github.com/deepfakes/faceswap}.
\newblock Accessed: 2020-06-27.

\bibitem{FOMMGit}
First order motion model official code.
\newblock \url{https://github.com/AliaksandrSiarohin/first-order-model}.
\newblock Accessed: 2020-07-16.

\bibitem{FSGANGit}
Fsgan official implementation.
\newblock \url{https://github.com/YuvalNirkin/fsgan}.
\newblock Accessed: 2020-07-16.

\bibitem{Avatarify}
Open avatarify.
\newblock \url{https://github.com/alievk/avatarify}.
\newblock Accessed: 2020-09-20.

\bibitem{Reface}
Reface.
\newblock \url{https://reface.ai/}.
\newblock Accessed: 2020-10-12.

\bibitem{selim}
{Selim Seferbekov's} dfdc winning model.
\newblock \url{https://github.com/selimsef/dfdc_deepfake_challenge}.
\newblock Accessed: 2020-09-24.

\bibitem{Wav2LipGit}
Wav2lip official code.
\newblock \url{https://github.com/Rudrabha/Wav2Lip}.
\newblock Accessed: 2020-08-24.

\bibitem{Agarwal01}
Shruti Agarwal, Hany Farid, Yuming Gu, Mingming He, Koki Nagano, and Hao Li.
\newblock Protecting world leaders against deep fakes.
\newblock {\em IEEE Conference on Computer Vision and Pattern Recognition
  (CVPR) Workshops}, pages 38--45, 2019.

\bibitem{Amerini01}
Irene Amerini, Leonardo Galteri, Roberto Caldelli, and Alberto~Del Bimbo.
\newblock Deepfake video detection through optical flow based cnn.
\newblock {\em IEEE International Conference on Computer Vision Workshop
  (ICCVW)}, pages 1205--1207, 2019.

\bibitem{Disinform2}
Dan Boneh, Andrew Grotto, Patrick McDaniel, and Nicolas Papernot.
\newblock Preparing for the age of deepfakes and disinformation.
\newblock {\em HAI Policy Brief}, 2020.

\bibitem{Bulat01}
Adrian Bulat and Georgios Tzimiropoulos.
\newblock How far are we from solving the 2d \& 3d face alignment problem? (and
  a dataset of 230,000 3d facial landmarks).
\newblock {\em IEEE International Conference on Computer Vision (ICCV)}, pages
  1021--1030, 2017.

\bibitem{Defame1}
Matt Burgess.
\newblock Deepfake porn is now mainstream. and major sites are cashing in.
\newblock {\em Wired}, 2020.

\bibitem{Chung02}
Joon~Son Chung, Andrew Senior, Oriol Vinyals, and Andrew Zisserman.
\newblock Lip reading sentences in the wild.
\newblock {\em IEEE Conference on Computer Vision and Pattern Recognition
  (CVPR)}, pages 3444--3453, 2017.

\bibitem{Chung01}
Joon~Son Chung and Andrew Zisserman.
\newblock Out of time: Automated lip sync in the wild.
\newblock {\em Computer Vision -- ACCV 2016 Workshops}, pages 251--263, 2017.

\bibitem{Dataset_DFDC}
Brian Dolhansky, Joanna Bitton, Ben Pflaum, Jikuo Lu, Russ Howes, Menglin Wang,
  and Cristian Ferrer.
\newblock The deepfake detection challenge (dfdc) dataset.
\newblock {\em arXiv preprint arXiv:2006.07397}, 2020.

\bibitem{Dataset_DFDCp}
Brian Dolhansky, Russ Howes, Ben Pflaum, Nicole Baram, and Cristian Ferrer.
\newblock The deepfake detection challenge (dfdc) preview dataset.
\newblock {\em arXiv preprint arXiv:1910.08854}, 2019.

\bibitem{Dataset_GDFD}
Nick Dufour and Andrew Gully.
\newblock Contributing data to deepfake detection research.
\newblock {\em Google AI Blog}, 2019.

\bibitem{Dataset_VFHQ}
Gereon Fox, Wentao Liu, Hyeongwoo Kim, Hans-Peter Seidel, Mohamed Elgharib, and
  Christian Theobalt.
\newblock Videoforensicshq: Detecting high-quality manipulated face videos.
\newblock {\em arXiv preprint arXiv:2005.10360}, 2020.

\bibitem{AdversarialGandhi}
Apurva Gandhi and Shomik Jain.
\newblock Adversarial perturbations fool deepfake detectors.
\newblock {\em 2020 International Joint Conference on Neural Networks (IJCNN)},
  pages 1--8, 2020.

\bibitem{AdversarialGaneshan}
Aditya Ganeshan, Vivek B.S., and R.~Venkatesh Babu.
\newblock Fda: Feature disruptive attack.
\newblock {\em IEEE International Conference on Computer Vision (ICCV)}, pages
  8068--8078, 2019.

\bibitem{AdversarialGoodfellow}
Ian~J. Goodfellow, Jonathon Shlens, and Christian Szegedy.
\newblock Explaining and harnessing adversarial examples.
\newblock {\em arXiv preprint arXiv:1412.6572}, 2015.

\bibitem{Goodman01}
Dou Goodman, Hao Xin, Wang Yang, Wu Yuesheng, Xiong Junfeng, and Zhang Huan.
\newblock Advbox: A toolbox to generate adversarial examples that fool neural
  networks.
\newblock {\em arXiv preprint arXiv:2001.05574}, 2020.

\bibitem{Defame2}
Karen Hao.
\newblock Deepfake porn is ruining women’s lives. now the law may finally ban
  it.
\newblock {\em MIT Technology Review}, 2021.

\bibitem{Hernandez-Ortega01}
Javier Hernandez-Ortega, Ruben Tolosana, Julian Fierrez, and Aythami Morales.
\newblock Deepfakeson-phys: Deepfakes detection based on heart rate estimation.
\newblock {\em Proc. 35th AAAI Conference on Artificial Intelligence
  Workshops}, 2021.

\bibitem{FID01}
Martin Heusel, Hubert Ramsauer, Thomas Unterthiner, Bernhard Nessler, and Sepp
  Hochreiter.
\newblock Gans trained by a two time-scale update rule converge to a local nash
  equilibrium.
\newblock {\em Advances in Neural Information Processing Systems (NeurIPS)},
  pages 6629--6640, 2017.

\bibitem{Huang01}
Yihao Huang, Felix Juefei-Xu, Qing Guo, Xiaofei Xie, Lei Ma, Weikai Miao, Yang
  Liu, and Geguang Pu.
\newblock Fakeretouch: Evading deepfakes detection via the guidance of
  deliberate noise.
\newblock {\em arXiv preprint arXiv:2009.09213}, 2020.

\bibitem{Hussain01}
Shehzeen Hussain, Paarth Neekhara, Malhar Jere, Farinaz Koushanfar, and Julian
  McAuley.
\newblock Adversarial deepfakes: Evaluating vulnerability of deepfake detectors
  to adversarial examples.
\newblock {\em IEEE Winter Conference on Applications of Computer Vision
  (WACV)}, pages 3348--3357, 2021.

\bibitem{Dataset_DF1.0}
Liming Jiang, Wayne Wu, Ren Li, Chen Qian, and Chen~Change Loy.
\newblock Deeperforensics-1.0: A large-scale dataset for real-world face
  forgery detection.
\newblock {\em IEEE Conference on Computer Vision and Pattern Recognition
  (CVPR)}, pages 2886--2895, 2020.

\bibitem{Dataset_DFTIMIT}
Pavel Korshunov and Sebastien Marcel.
\newblock Deepfakes: a new threat to face recognition? assessment and
  detection.
\newblock {\em arXiv preprint arXiv:1812.08685}, 2018.

\bibitem{Li01}
Xiaodan Li, Yining Lang, Yuefeng Chen, Xiaofeng Mao, Yuan He, Shuhui Wang, Hui
  Xue, and Quan Lu.
\newblock Sharp multiple instance learning for deepfake video detection.
\newblock {\em MM '20: Proceedings of the 28th ACM International Conference on
  Multimedia}, pages 1864–--1872, 2020.

\bibitem{Dataset_CDF}
Yuezun Li, Xin Yang, Pu Sun, Honggang Qi, and Siwei Lyu.
\newblock Celeb-df: A large-scale challenging dataset for deepfake forensics.
\newblock {\em IEEE Conference on Computer Vision and Pattern Recognition
  (CVPR)}, pages 3204--3213, 2020.

\bibitem{Masi01}
Iacopo Masi, Aditya Killekar, Royston Mascarenhas, Shenoy Gurudatt, and Wael
  AbdAlmageed.
\newblock Two-branch recurrent network for isolating deepfakes in videos.
\newblock {\em Computer Vision -- ECCV 2020)}, pages 667--684, 2020.

\bibitem{Nagrani01}
Arsha Nagrani, Joon~Son Chung, and Andrew Zisserman.
\newblock Voxceleb: A large-scale speaker identification dataset.
\newblock {\em Proc. Interspeech 2017}, pages 2616--2620, 2017.

\bibitem{Neekhara01}
Paarth Neekhara, Brian Dolhansky, Joanna Bitton, and Cristian Ferrer.
\newblock Adversarial threats to deepfake detection: A practical perspective.
\newblock {\em arXiv preprint arXiv:2011.09957}, 2020.

\bibitem{Models_FSGAN}
Yuval Nirkin, Yosi Keller, and Tal Hassner.
\newblock Fsgan: Subject agnostic face swapping and reenactment.
\newblock {\em IEEE International Conference on Computer Vision (ICCV)}, pages
  7184--7193, 2019.

\bibitem{KoreanDict}
National~Institute of Korean~Language.
\newblock Standard korean language dictionary.
\newblock \url{https://stdict.korean.go.kr/main/main.do}.
\newblock Accessed: 2020-06-25.

\bibitem{KunsanSentiment}
SangMin Park, ChulWon Na, MinSeong Cho, DaHee Lee, and ByungWon On.
\newblock Knu korean sentiment lexicon: Bi-lstm-based method for building a
  korean sentiment lexicon.
\newblock {\em Journal of Intelligence and Information Systems},
  24(4):219--240, 2018.

\bibitem{Models_DFL}
Ivan Petrov, Daiheng Gao, Nikolay Chervoniy, Kunlin Liu, Sugasa Marangonda,
  Chris Um'e, Mr. dpfks, RP Luis, Jian Jiang, Sheng Zhang, Pingyu Wu, Bo Zhou,
  and Weiming Zhang.
\newblock Deepfacelab: A simple, flexible and extensible face swapping
  framework.
\newblock {\em arXiv preprint arXiv:2005.05535}, 2020.

\bibitem{Models_Wav2Lip}
KR Prajwal, Rudrabha Mukhopadhyay, Vinay Namboodiri, and CV Jawahar.
\newblock A lip sync expert is all you need for speech to lip generation in the
  wild.
\newblock {\em MM '20: Proceedings of the 28th ACM International Conference on
  Multimedia}, 2020.

\bibitem{Dataset_FFPP}
Andreas R{\"o}ssler, Davide Cozzolino, Christian~Riess Luisa~Verdoliva, Justus
  Thies, and Matthias Nie{\ss}ner.
\newblock Faceforensics++: Learning to detect manipulated facial images.
\newblock {\em IEEE International Conference on Computer Vision (ICCV)}, pages
  1--11, 2019.

\bibitem{Sabir01}
Ekraam Sabir, Jiaxin Cheng, Ayush Jaiswal, Wael AbdAlmageed, Iacopo Masi, and
  Prem Natarajan.
\newblock Recurrent convolutional strategies for face manipulation detection in
  videos.
\newblock {\em IEEE Conference on Computer Vision and Pattern Recognition
  (CVPR) Workshops}, pages 80--87, 2019.

\bibitem{Models_FOMM}
Aliaksandr Siarohin, St{\'e}phane Lathuili{\`e}re, Sergey Tulyakov, Elisa
  Ricci, and Nicu Sebe.
\newblock First order motion model for image animation.
\newblock {\em Advances in Neural Information Processing Systems (NeurIPS)},
  2019.

\bibitem{Fraud1}
Catherine Stupp.
\newblock Fraudsters used ai to mimic ceo’s voice in unusual cybercrime case.
\newblock {\em The Wall Street Journal}, 2019.

\bibitem{pmlr-v97-tan19a}
Mingxing Tan and Quoc Le.
\newblock Efficientnet: Rethinking model scaling for convolutional neural
  networks.
\newblock {\em Proceedings of the 36th International Conference on Machine
  Learning}, 97:6105--6114, 2019.

\bibitem{Tolosana01}
Ruben Tolosana, Sergio Romero-Tapiador, Julian Fierrez, and Ruben
  Vera-Rodriguez.
\newblock Deepfakes evolution: Analysis of facial regions and fake detection
  performance.
\newblock {\em Pattern Recognition. ICPR International Workshops and
  Challenges}, pages 442--456, 2021.

\bibitem{Tolosana02}
Ruben Tolosana, Ruben Vera-Rodriguez, Julian Fierrez, Aythami Morales, and
  Javier Ortega-Garcia.
\newblock Deepfakes and beyond: A survey of face manipulation and fake
  detection.
\newblock {\em Information Fusion}, 64:131--148, 2020.

\bibitem{Disinform1}
Cristian Vaccari and Andrew Chadwick.
\newblock Deepfakes and disinformation: Exploring the impact of synthetic
  political video on deception, uncertainty, and trust in news.
\newblock {\em Social Media + Society}, pages 1--13, 2020.

\bibitem{Fraud2}
Jonathan Vanian.
\newblock Why american express is trying technology that makes deepfake videos
  look real.
\newblock {\em Fortune}, 2020.

\bibitem{Wang01}
Sheng-Yu Wang, Oliver Wang, Richard Zhang, Andrew Owens, and Alexei~A. Efros.
\newblock Cnn-generated images are surprisingly easy to spot… for now.
\newblock {\em IEEE Conference on Computer Vision and Pattern Recognition
  (CVPR)}, pages 8692--8701, 2020.

\bibitem{Dataset_UADFV}
Xin Yang, Yuezun Li, and Siwei Lyu.
\newblock Exposing deep fakes using inconsistent head poses.
\newblock {\em IEEE International Conference on Acoustics, Speech and Signal
  Processing (ICASSP)}, 2019.

\bibitem{Models_ATFHP}
Ran Yi, Zipeng Ye, Juyong Zhang, Hujun Bao, and Yongjin Liu.
\newblock Audio-driven talking face video generation with learning-based
  personalized head pose.
\newblock {\em arXiv preprint arXiv:2002.10137}, 2020.

\bibitem{MTCNN}
Kaipeng Zhang, Zhanpeng Zhang, Zifeng Li, and Yu Qiao.
\newblock Joint face detection and alignment using multitask cascaded
  convolutional networks.
\newblock {\em IEEE Signal Processing Letters}, 23(10):1499--1503, 2016.

\end{thebibliography}
}

\end{document}